\documentclass[runningheads]{llncs}
\usepackage{graphicx}
\usepackage{amsmath,amssymb} 
\usepackage{acronym}
\usepackage{caption}
\usepackage{subfig}
\usepackage{color}
\usepackage{multirow}

\graphicspath{{Images/}}

\acrodef{VSLAM}{Visual Simultaneous Localisation And Mapping}
\acrodef{MCL}{Monte Carlo Localisation}
\acrodef{FCN}{Fully Convolutional Network}
\acrodef{FCU-Net}{Fully Convolutional U-Net}
\acrodef{SONAR}{SOund Navigation And Ranging}
\acrodef{LiDAR}{Light Detection And Ranging}
\acrodef{LK}{Lucas-Kanade}
\acrodef{VO}{Visual Odometry}

\mathchardef\mhyphen="2D
\DeclareMathOperator*{\argmin}{arg\,min}
\DeclareMathOperator*{\argmax}{arg\,max}

\begin{document}

\pagestyle{headings}
\mainmatter
\def\ECCV18SubNumber{3}

\title{Localisation via Deep Imagination: learn the features not the map\vspace{-0.7cm}} 
\titlerunning{Localisation via Deep Imagination}
\authorrunning{J. Spencer, O. Mendez, R. Bowden, S. Hadfield}

\author{Jaime Spencer \and Oscar Mendez \and Richard Bowden \and Simon Hadfield\\ \footnotesize{\email{\{jaime.spencer, o.mendez, r.bowden, s.hadfield\}@surrey.ac.uk}}\vspace{-0.2cm}}

\institute{University of Surrey\vspace{0.2cm}}

\maketitle
\begin{abstract}
\vspace{-1cm}
How many times does a human have to drive through the same area to become familiar with it? 
To begin with, we might first build a mental model of our surroundings. 
Upon revisiting this area, we can use this model to extrapolate to new unseen locations and imagine their appearance.

Based on this, we propose an approach where an agent is capable of modelling new environments after a single visitation. 
To this end, we introduce ``Deep Imagination'', a combination of classical Visual-based Monte Carlo Localisation and deep learning. 
By making use of a feature embedded 3D map, the system can ``imagine'' the view from any novel location. 
These ``imagined'' views are contrasted with the current observation in order to estimate the agent's current location. 
In order to build the embedded map, we train a deep Siamese Fully Convolutional U-Net to perform dense feature extraction. 
By training these features to be generic, no additional training or fine tuning is required to adapt to new environments.

Our results demonstrate the generality and transfer capability of our learnt dense features by training and evaluating on multiple datasets. 
Additionally, we include several visualizations of the feature representations and resulting 3D maps, as well as their application to localisation.

\vspace{-0.4cm}
\end{abstract}

\vspace{-1cm}
\section{Introduction}
\vspace{-0.4cm}
Localisation is fundamental to interacting with the world. 
Previous knowledge of an existing environment can greatly improve localisation accuracy. 
Despite this, localisation is still especially challenging in highly dynamic environments such as vehicle automation, where independently moving distractor objects make online localisation difficult, and rapid reactions are needed.
Typically, localisation approaches have relied on expensive and power hungry sensors such as \ac{LiDAR}. 
This is not feasible if these systems are to be made available to the general consumer public.

In order to reduce sensor cost, \ac{VSLAM} algorithms have been proposed, where both the scenery map and agent's locations are estimated during test time. 
This can help simplify the problem and reduce external dependencies. 
However, \ac{VSLAM} tends to suffer from reduced accuracy due to it's susceptibility to drift. 
In contrast, another solution is to separate the data collection and training from the deployment stage. 
During training, a single vehicle equipped with the necessary sensors can collect the required data and maps. 
During deployment, other agents can exploit this data and solve for localisation in a purely visual manner using a low-cost RGB camera.\looseness=-1

\ac{MCL} is considered the state-of-the-art in many applications. 
However, traditional implementations require \ac{SONAR} and \ac{LiDAR}, making them quite expensive. 
More recent work has focused on the use of visual information coupled with additional sensors such as RGB-D, GPS or IMUs. 
These methods suffer due to unreliable tracking of visual features, caused by environmental appearance changes. 
Additionally, drift and error accumulation can be hard to detect and correct, leading to large errors. 
With the advent of deep learning, solutions making use of end-to-end pose regression networks have gained popularity. 
Still, these suffer from scalability issues since the networks must be retrained or fine tuned to each new environment.\looseness=-1

Instead, we propose a biologically inspired approach. 
Humans can perform global localisation using prebuilt representations of the world, including maps, floorplans and 3D models. 
By imagining the appearance of the world from different locations and comparing it to our own observations, we can estimate our position in the given representation.
Based on this, we aim to solve the localisation problem by providing the system with ``Deep Imagination''. 
By making use of a deep dense feature extractor, a feature embedded 3D world map is created. 
Each point in the map is associated with an \textit{n}-dimensional feature descriptor in addition to it's \textit{xyz} coordinates. 
These features are trained to be invariant to changes in appearance. 
The system can now use this enriched map to ``imagine'' the view from any given position. 
By contrasting candidate ``imagined'' views to the actual observation, a likelihood for each position can be obtained. 
If the feature extractor is trained to be generic, little to no training data is required to adapt to unseen environments. 
In turn, this means that our system only requires one visit to build the required representation of any new scene. 
From this representation, the agent can now ``imagine'' the view from any new viewpoint.\looseness=-1

One of the challenging aspects faced by map learning systems is the constant appearance change of the environment. 
Some of these changes can be gradual, such as those dependent on the time of day, seasons and weather, whereas others are more dynamic, such as occlusions, lighting variations or unique vehicles and pedestrians at varying locations. 
This demands a level of feature invariance and robustness that allows point matching regardless of current appearance.

To counteract this during map generation time, we propose using a deep Siamese \ac{FCU-Net} to extract dense features from each image, which are backprojected into the 3D world. 
By following this approach, training is restricted exclusively to the feature extraction. 
If these features are pretrained to be generic, no additional training is required to extend the system to new locations. 
In turn, this means that our approach is much more scalable and easy to adapt than map learning methods.

The rest of the paper is structured as follows. 
Section \ref{sec: lit} introduces previous solutions to localisation, including \ac{MCL} and deep learning approaches. 
The details and implementation of our system can be found in Section \ref{sec: meth}. 
This includes the feature extraction and map building, along with the ``Deep Imagination'' localiser and \ac{VO} motion estimator. 
Section \ref{sec: res} presents various experiments used to validate our method, including the datasets used, training regime, generality and transfer capability of the learnt descriptors and localisation performance. 
Finally, conclusions and future work can be found in Section \ref{sec: conc}.\looseness=-1

\vspace{-0.6cm}
\section{Literature Review} 
\label{sec: lit}
\vspace{-0.4cm}
Early localisation methods employed Kalman filtering or grid-based Markov approaches. 
Monte Carlo Localisation (MCL), introduced by Fox \textit{et al.} \cite{Fox1999} and Dellaert \textit{et al.} \cite{Dellaert1999}, built on these methods by representing the location probability distribution function with particles randomly sampled from the distribution. 
The arrival of sensors such as \ac{SONAR} and \ac{LiDAR} allowed \ac{MCL} to become the state-of-the-art approach to accurate localisation. 
However, these sensors use cumbersome sonar arrays and range finders, making them expensive and impractical. 
These methods are known as Range-based \ac{MCL}.

\ac{MCL} has since been adapted to use various kinds of sensors, most notably visual ones \cite{Dellaert1999_2}. 
This gave rise to Vision-based \ac{MCL} (VMCL), allowing for the use of cheaper sensors at the cost of reduced robustness. 
The accuracy of these methods was improved through invariant feature extractors (SIFT, Gist, Harris) \cite{ParraAlonso2012,Li2013} and the addition of complementary sensors (RGB-D, GPS, IMU) \cite{Wei2011,Gao2015}. 
Paton and Kosecka \cite{Paton2012} combine SIFT feature matching and ICP. 
Kamijo \textit{et al.} \cite{KAMIJO2015} combine GPS/IMU global positioning with visual lane markers for lateral positioning. 
More recently, semantic information has been used instead of additional sensors via SEmantic Detection and Ranging (SeDAR) \cite{Mendez2018}.

Non-\ac{MCL} approaches commonly learn a representation of a fixed map from which camera position can be regressed. 
Shotton \textit{et al.} \cite{Shotton2013} train a regression forest to predict dense correspondences on raw RGB-D images, removing the need for feature extractors. 
Kendall \textit{et al.} \cite{Kendall,Kendalla} instead opt for an end-to-end deep learning approach, PoseNet, which uses transfer learning to adapt the network to new scenes. 
Melekhov \textit{et al.} \cite{Melekhov2018} improve on PoseNet by using an hourglass network with skip connections. 
Brachmann \textit{et al.} \cite{Brachmann} introduce a differentiable version of RANSAC (DSAC), which can be included within the training pipeline of a network.
RNNs and LSTMs have also recently adapted to this line of work in order to take advantage of temporal information \cite{Clark2017,Walch}. 

Other approaches, typically associated with \ac{VSLAM}, focus on the map production. 
In this context, a set of 3D landmarks and their associated descriptors are defined as maps using Bayesian filtering \cite{Davison2003}, key-frame methods \cite{Mur-Artal2015} or bundle adjustment \cite{Mouragnon2006,Eudes}. 
Mendez \textit{et al.} \cite{Mendez2017} \cite{Mendez} focus on environment reconstruction using collaborative agents.
Wang \textit{et al.} \cite{Wang2018a} build a 3D semantic map of the world and use GPU/IMU sensor fusion to refine pose. 
On the other hand, Brahmbhatt \textit{et al.} \cite{Brahmbhatt2017} aim to learn a general map representation as weights of a DNN trained to regress camera pose. 

We propose a combination of VMCL and deep learning. 
In a similar fashion to \cite{Wang2018a}, we build an augmented 3D map. 
However, coarse semantic labels are replaced with dense and invariant feature representations. 
Location is then obtained via ``Deep Imagination''. 
Expected views are combined with the current observation in a novel VMCL particle filter. 
Since the dense extracted features can be pretrained to be generic, a new embedded 3D map can be generated from a single run through the training data for a new scene. 
This greatly enhances the scalability of our system. 

\section{Methodology}
\label{sec: meth}
\vspace{-0.5cm}
Contrary to most MCL methods, we propose a fully visual system without reliance on additional range-based sensors. 
By using generic features, we limit training to a single Siamese \ac{FCU-Net} that can be used in multiple environments. 
To adapt the system to a new world, we can simply obtain the feature representation of each available image and backproject it onto the built map. 

\begin{figure}[tb!]
\centering
\includegraphics[width=0.8\textwidth]{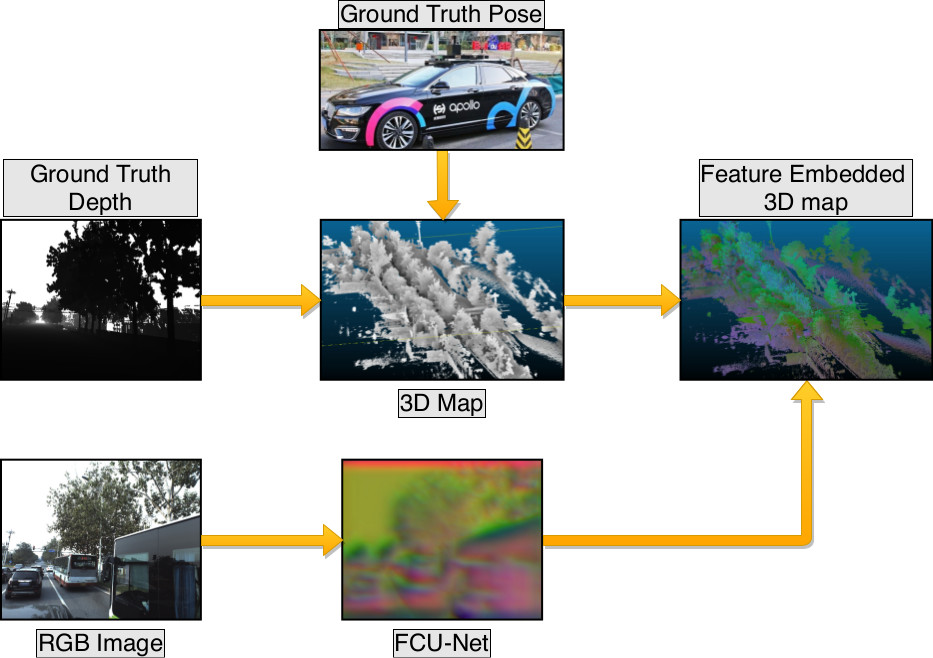}
\vspace{-0.2cm}
\caption{Map building overview. 3D location of a point is provided as part of the ground truth labels. A pretrained \ac{FCU-Net} performs the dense feature extraction and creates the embedded 3D map representation.}
\vspace{-0.1cm}
\label{fig:train_flow}
\end{figure}

The overview for the feature embedded 3D map generation can be found in Figure \ref{fig:train_flow}. 
From the ground truth pose and depth for each image, it's corresponding location can be found using simple projective geometry. 
The pretrained network is then used as a dense feature extractor and fused with the base map.   

\begin{figure}[tb!]
\centering
\vspace{-0.2cm}
\includegraphics[width=0.8\textwidth]{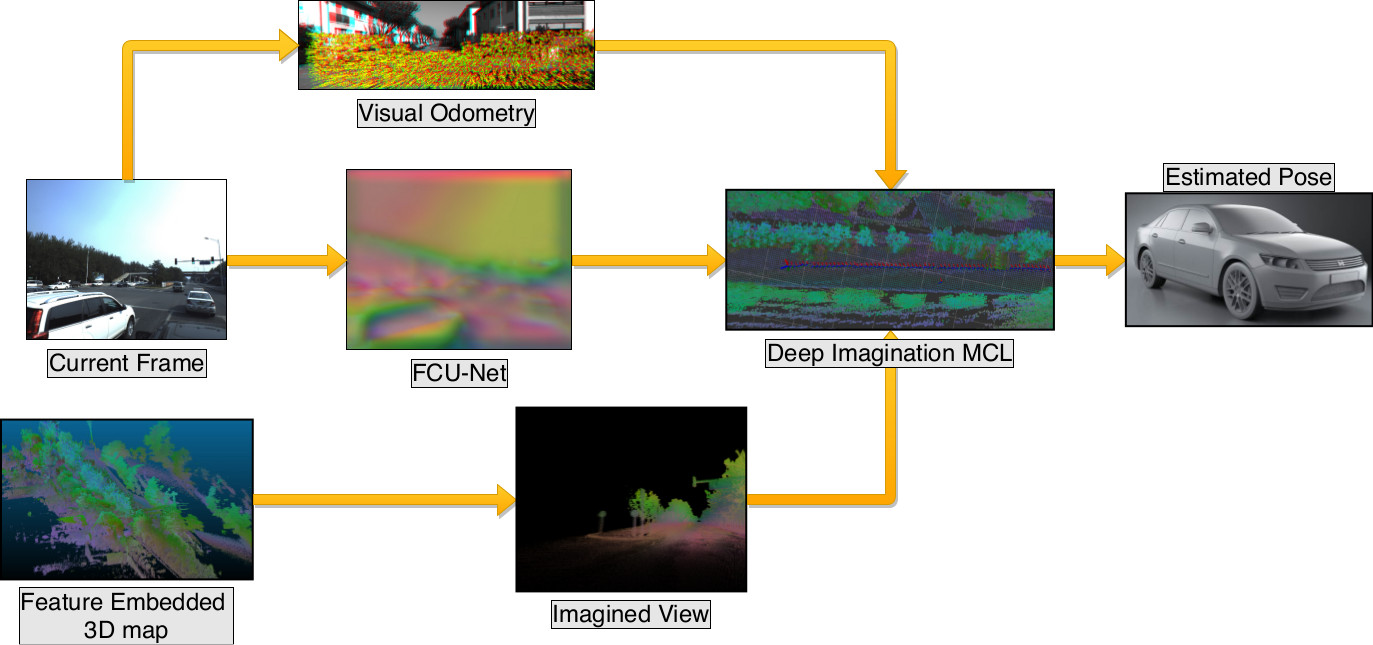}
\vspace{-0.1cm}
\caption{Deployment workflow overview. The current frame is used in \ac{VO} motion estimation. It's dense feature representation is then obtained from the pretrained \ac{FCU-Net}. Through ``Deep Imagination'', an estimated view for each candidate position is obtained.}
\vspace{-0.5cm}
\label{fig:test_flow}
\end{figure}

The deployment phase diagram is shown in Figure \ref{fig:test_flow}. 
The ``Deep Imagination'' localiser lies at the core of the implementation, making use of a VMCL particle filter. 
As seen, the localiser uses the feature embedded 3D map previously generated in Figure \ref{fig:train_flow}. 
Through the map, the system ``imagines'' what the world should look like from previously unseen viewpoints. 

At run time, the current and previous observations are used to estimate motion between frames with \ac{VO}. 
This is done via essential matrix estimation using matched features between the images. 
RANSAC provides an additional refinement step and robustness to outliers. 
This position, however, is only locally accurate and is susceptible to drift. 
The current dense feature representation is obtained from the pretrained \ac{FCU-Net}. 
Combined with the estimated \ac{VO} motion and ``imagined'' viewpoints, pose likelihoods are obtained and propagated using the VMCL particle filter within the ``Deep Imagination'' localiser.

\subsection{FCU-Net feature extraction}
\vspace{-0.2cm}
In order to efficiently train a general deep learning solution for dense feature extraction a Siamese \ac{FCU-Net} is employed. 
\acp{FCN} have previously been used for tasks such as pixel-wise segmentation \cite{Long2015}, where the output is required to have the same resolution as the input. 
By employing only convolutions, the network isn't restricted to a specific input size and maintains a relatively constant inference time, given that each pixel can be processed in parallel. 

\begin{figure}[b!]
\centering
\vspace{-0.6cm}
\includegraphics[width=0.6\textwidth]{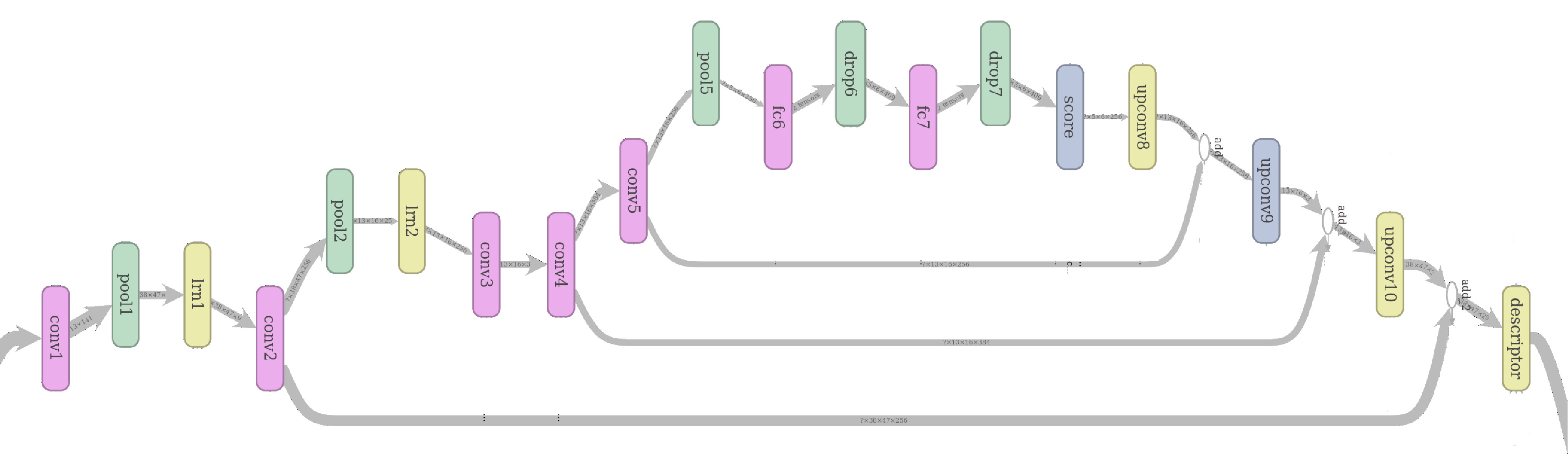}
\vspace{-0.2cm}
\caption{\ac{FCU-Net} branch for dense feature extraction. Skip connections connecting corresponding sized layers improve the spatial resolution of the final descriptor. Employing an \ac{FCN} model allows for variable sized image inputs.}
\label{fig: fcn}
\vspace{-0.4cm}
\end{figure}

The architecture used in the \ac{FCU-Net} is shown in Figure \ref{fig: fcn}. 
It consists of a downsampling stage (encoder), followed by a bottleneck layer and an upsampling stage (decoder). 
Layers \textit{fc6} and \textit{fc7} replace traditional fully connected layers with 1x1 convolutions. 
To improve the spatial resolution of the output, skip connections between corresponding sized layers in both stages are added. 
This allows for the combination of low level spatial information from earlier layers with higher level semantic meaning from deeper layers. 

Given an input image $I$, it's dense feature representation can be obtained by 
\begin{equation}
	F(\boldsymbol{p}) = U(I(\boldsymbol{p}) | w),
\end{equation}
where $\boldsymbol{p}$ represents a 2D point and $U$ represents an \ac{FCU-Net}, parametrized by a set of weights $w$. 
$I$ stores an RGB colour value, whereas $F$ stores an $n$-dimensional feature descriptor, $U: \mathbb{N}^3 \to \mathbb{R}^n$.

We build on the ideas presented in \cite{Schmidt2017} and propose a ``pixel-wise" contrastive loss \cite{Chopra2005a}. 
A Siamese network with two identical \ac{FCU-Net} branches is trained using pixel-wise contrastive loss to produce dense descriptor maps. 
Given a pair of input points, contrastive loss is defined as
\begin{equation} \label{eq: contrastive}
	l(y, \boldsymbol{p}_1, \boldsymbol{p}_2) = 
    \begin{cases}
    \frac{1}{2}(d)^2 & \text{if } y = 1 \\
    \frac{1}{2}\{\max(0, m - d)\}^2 & \text{if } y = 0 \\
    0 & otherwise
    \end{cases}
\end{equation}
where $d$ is the euclidean distance of the feature embeddings $||F_1(\boldsymbol{p}_1) - F_2(\boldsymbol{p}_2)||$, and $y$ is the label indicating if the pair is a match and $m$ is the margin. 
Intuitively, similar points (matching pairs) should be close in the latent space, while dissimilar points (non-matching pairs) should be separated by at least the margin.\looseness=-1

We can use this loss to learn relocalisation features by projecting homogeneous 3D world points, $\dot{\boldsymbol{q}}$, onto pairs of images. 
A set of corresponding pixels can be obtained through
\begin{equation}
\boldsymbol{p} = \pi(\dot{\boldsymbol{q}}|\boldsymbol{K}, \boldsymbol{P}) = \boldsymbol{K} \boldsymbol{P} \dot{\boldsymbol{q}},
\end{equation}
\begin{equation}
\label{eq: proj}
\pi(\dot{\boldsymbol{q}}|\boldsymbol{K}_1, \boldsymbol{P}_1 ) \mapsto \pi(\dot{\boldsymbol{q}}|\boldsymbol{K}_2, \boldsymbol{P}_2),
\end{equation}
where $\pi$ is the projection function and $\boldsymbol{K}$ and $\boldsymbol{P}$ represent the camera's intrinsics and global pose, respectively.

From these points, a label mask $\boldsymbol{Y}$ is created, indicating if each pair of pixels is a match, non-match or should be ignored. 
Unlike a traditional Siamese network, every input image has many matches, which are not spatially aligned. 
As an extension to (\ref{eq: contrastive}) we obtain
\begin{equation}
	L(\boldsymbol{Y}, \boldsymbol{F}_1, \boldsymbol{F}_2) = \sum_{i \in \boldsymbol{p}_1} \sum_{j \in \boldsymbol{p}_2} l(\boldsymbol{Y}(i, j), \boldsymbol{F}_1(i), \boldsymbol{F}_2(j)).
\end{equation}
\vspace{-0.6cm}
\subsection{Feature embedded 3D map}
\vspace{-0.2cm}
Once the network has been pretrained, the feature embedded map can be built. 
This map is needed to support the ``Deep Imagination'' localiser. 
During map building, each stereo RGB pair has an associated depth image ($\boldsymbol{D}$) and 3D pose ($\boldsymbol{P}$), along with camera calibration parameters. 
The 3D location of any given pixel can be obtained by
\begin{equation}
	\boldsymbol{q} = \pi^{-1}(\dot{\boldsymbol{p}}|\boldsymbol{K}, \boldsymbol{P}) = \boldsymbol{K}^{-1}  \boldsymbol{P}^{-1} \dot{\boldsymbol{p}} D(\boldsymbol{p}),
\end{equation}
where $\pi^{-1}$ is the backprojection function.

Additionally, dense features are extracted from each frame and associated with their corresponding 3D location in an octomap $\mathcal{M}$. 
Since most voxels are visible in more than one frame, there will be multiple available descriptors. 
To reduce memory requirements and produce a more robust representation, the stored descriptor is the average of all \textit{k} observations of that voxel 
\begin{equation}
	\mathcal{M}(\boldsymbol{q}) = \frac{1}{k} \sum_{i \in k} F_i(\boldsymbol{p}) \text{ where } \pi^{-1}(\dot{\boldsymbol{p}}|\boldsymbol{K}_i, \boldsymbol{P}_i) = \boldsymbol{q}.
\end{equation}

\vspace{-0.5cm}
\subsection{Deep imagination MCL}
\label{sec: dil}
\vspace{-0.2cm}
The \ac{VO} system, described in Section \ref{sec: vo}, provides an invaluable source of information, which allows us to efficiently estimate incremental location updates.
However, the iterative nature of it's estimate lead to two major problems.
Firstly, with no fixed point of reference, an agent can only describe it's location relative to it's starting point. 
Being unable to understand it's location in absolute terms makes it impossible to cooperate with other intelligent agents, or to exploit the prior environmental knowledge generated above.
Secondly, the accumulation of errors at every incremental estimation, will invariably lead to drift over time, providing a hard limit on the reliable operational time of the system.
To resolve both of these issues, it is necessary to incorporate a global localizer. 
This localizer provides independent, non-incremental estimates of location (to eliminate drift) while also anchoring the agent's pose within an absolute co-ordinate frame that can be shared with other agents.

Humans perform this absolute global localization using maps, floorplans, 3D models, or other prebuilt representations of their environment. 
Generally, a person will examine this map, and try to imagine what the world it represents would look like from different locations. 
By contrasting these imagined views against their real-world observations, we are able to eventually determine our location in the co-ordinate frame of the map.
Inspired by this, our approach attempts to imagine what different environmental viewpoints would look like according to a deep feature embedding network. 
By correlating the embedded representation of our observations, against the imagined embedding, we can determine our location in a way that is robust to lighting and environmental factors.

We define a compact representation of a pose ($\boldsymbol{P}$) with 6 degrees of freedom, as $\boldsymbol{\boldsymbol{\omega}} \in \text{SE}(3)$. 
This comprises 3 translational degrees of freedom and 3 rotational degrees of freedom. 
The probability of a particular pose at time $t$, conditioned on a series of observations can then be defined as $P_{1..t}(\boldsymbol{\boldsymbol{\omega}}|I_{1..t},\mathcal{M})$. 
We adopt the standard iterative Bayesian filtering formulation
\begin{equation}
P_{1..t}(\boldsymbol{\omega}|I_{1..t},\mathcal{M}) =
P_{t}(I_{t}|\boldsymbol{\omega}_{t},\mathcal{M})\,\,
P(\boldsymbol{\omega}_{t} | \boldsymbol{\omega}_{t\mhyphen 1})\,\,
P_{1..t\mhyphen 1}(\boldsymbol{\omega}|I_{1..t\mhyphen 1},\mathcal{M}),
\label{eq:post}
\end{equation}
where $P(\boldsymbol{\omega}_{t} | \boldsymbol{\omega}_{t\mhyphen 1})$ is the transition function and the likelihood measurements are assumed to be independent at each time step.
Thus, to compute the probability of any pose we need only know the likelihood of that pose, the transition function, and the probability distribution at the previous frame.
We can compute the likelihood of any given pose by first imagining the deep feature embedding $\hat{F}_{\boldsymbol{\omega}}$ of the world from that perspective. 
This is done by projecting the feature embedded map to the candidate pose such that
\begin{equation}
\begin{aligned}
\hat{F}_{\boldsymbol{\omega}}(\boldsymbol{p}) = \mathcal{M}(\boldsymbol{q}) & & \mathrm{ where } & & \pi(\dot{\boldsymbol{q}}|\boldsymbol{K},\boldsymbol{\omega})=\boldsymbol{p}.
\end{aligned}
\end{equation}

To deal with occlusions we add the further constraint that 
\begin{equation}
\boldsymbol{q} = \argmin_{\bar{\boldsymbol{q}} \in \mathcal{M}}D\left(\bar{\boldsymbol{q}}, \boldsymbol{\omega}\right),
\end{equation}
where $D$ computes the euclidean distance between the voxel and the hypothesis pose.
We can now define the pose likelihood by contrasting this imagined deep embedding against the true embedding of our observations
\begin{equation}
P_{t}(I_{t}|\boldsymbol{\omega}_{t},\mathcal{M}) = \exp\left(-\sigma_{l}\left|\hat{F}_{\boldsymbol{\omega}}-F_{t}\right|_{1}\right),
\end{equation}
where $\sigma_{l}$ is a scaling factor which is inversely proportional to the number of entries in the feature embedding.
Finally, we can exploit the relative motion $\Delta\boldsymbol{\omega}$ estimated by the visual odometry system (Section~\ref{sec: vo}) to define our transition function as
\begin{equation}
P(\boldsymbol{\omega}_{t} | \boldsymbol{\omega}_{t\mhyphen 1}) = P(\boldsymbol{\omega}_{t\mhyphen 1}) + \mathcal{N}(\Delta\boldsymbol{\omega},\boldsymbol{\Sigma}_o),
\label{eq:trans}
\end{equation}
where $\boldsymbol{\Sigma}_o$ is the covariance matrix modelling the uncertainty characteristics of the visual odometry system. 

We now have a complete definition for the posterior probability of any pose, given a series of observations.
For efficiency, we approximate this distribution with a collection of samples $S=\{\boldsymbol{s}_1,..,\boldsymbol{s}_N\}$ with associated weights $w_1,..,w_N$.

To produce the final estimate of the location, we first run a weighted mean-shift clustering on the samples which are approximating the posterior distribution (\ref{eq:post}). 
The collection of cluster centres $\bar{S}=\{\bar{\boldsymbol{s}}_1,..,\bar{\boldsymbol{s}}_N\}$ is iteratively updated
\begin{equation}
\bar{\boldsymbol{s}}_{n} = \frac{
	\displaystyle\sum_{\boldsymbol{s}_{i} \in S} D_G(\bar{\boldsymbol{s}}_{n}, \boldsymbol{s}_{i}) \boldsymbol{s}_{i} w_{i}
}
{
	\displaystyle\sum_{\boldsymbol{s}_{i} \in S} D_G(\bar{\boldsymbol{s}}_{n}, \boldsymbol{s}_{i}) w_{i}
},
\end{equation}
where $D_G$ applies a Gaussian kernel on the distance between two samples. Once the clustering has converged, the final estimate of the location $\tilde{\boldsymbol{s}}$ is given by the centroid of the largest weighted cluster %
\begin{equation}
\tilde{\boldsymbol{s}} = \argmax_{\bar{\boldsymbol{s}} \in \bar{S}} \displaystyle\sum_{\boldsymbol{s}_{i} \in S} D_G(\bar{\boldsymbol{s}}_{n}, \boldsymbol{s}_{i}) w_{i}.
\end{equation}
This approximates a \textit{Maximum A Posteriori} estimate, encoding both the prior distribution of points, and their likelihood weightings.

\vspace{-0.3cm}
\subsection{Visual odometry}
\label{sec: vo}
\vspace{-0.2cm}
In order to extract local movement during test time, SIFT feature matching is performed on a pair of consecutive frames, $I_t$ \& $I_{t\mhyphen 1}$. 
From these matches, camera motion is obtained via essential matrix estimation. 
The essential matrix $\boldsymbol{E}$ $\in \text{SE}(3)$ represents the translation ($\boldsymbol{t}$) and rotation ($\boldsymbol{R}$) between two views,
\begin{equation}
	\boldsymbol{E} = [\boldsymbol{t}]_x \boldsymbol{R},
	\quad \textrm{ where } \quad
    \dot{\boldsymbol{p}}_1\boldsymbol{E}\dot{\boldsymbol{p}}_2 = 0.
\end{equation}
where $[\boldsymbol{t}]_x$ is the matrix-representation of the vector cross-product. 
$\boldsymbol{E}$ can be decomposed via Single Value Decomposition and estimated using a minimum of five point correspondences. 
However, there are four possible combinations of $\boldsymbol{R}$ and $\boldsymbol{t}$ that solve for $\boldsymbol{E}$. 
In order to determine the correct pair, a 3D reconstruction for each is performed.
The pair with the largest proportion of points in front of both cameras is selected as the correct one. 
By combining this with RANSAC, a more robust estimate in the presence of noise and outliers can be obtained. 

However, it is well understood that monocular odometry suffers from scale ambiguity. 
This is normally resolved using depth sensors. 
In our work, we assume the depth sensors are not present during revisitation at deployment time.
While there exist methods to recover the scale on a monocular system without a depth sensor, they are beyond the scope of this paper. 
Instead we exploit the non-holonomic nature of the vehicle and use a constant velocity motion model to scale the visual odometry measurements to the expected displacement, resulting in $\Delta\boldsymbol{\boldsymbol{\omega}} \in \text{SE}(3)$ as used in (\ref{eq:trans}).

\vspace{-0.4cm}
\section{Results}
\label{sec: res}
\vspace{-0.4cm}
We make use of the Kitti odometry dataset \cite{Geiger} to pretrain our Siamese \ac{FCU-Net}. 
The odometry dataset provides various sequences of stereo pairs, along with a corresponding rectified Velodyne pointcloud and camera locations. 
Only a subsection from sequence `00' (over 4500 stereo pairs) is used to train the network. 
Once a base pointcloud has been built from the available frames, it is projected onto pairs of images to obtain correspondence between them, as per (\ref{eq: proj}). 
A total of 664 pairs are used for training, while 174 are used for validation. 
Since each pair has approximately 13000 matches, this corresponds to $8.6\times10^6$ training examples. 
All networks are trained from scratch on this dataset, without any additional pretraining.

To test the generality of our learnt features and perform the final localisation, we use the Apollo Scape dataset \cite{Wang2018a}. 
Ground truth poses for each of the stereo pairs, along with a 3D semantic pointcloud is provided. 
Test videos are recorded in the same scene with different conditions, hence requiring invariant feature detection. 
From the multiple roads and sequences available, a subset of 562 pairs are used for training and 144 pairs for validation. 
Once again, each pair has an average of 15000 matches, giving a total of $8.4\times10^6$ training examples.

\vspace{-0.4cm}
\subsection{Siamese FCU-Net training}
\vspace{-0.3cm}
In order to train our feature descriptor network, we use the previously mentioned Siamese network consisting of two identical \ac{FCU-Net} branches. 
The whole system was implemented in TensorFlow \cite{Abadi2015}. 
Each training item consists of a pair of images and a set of correspondences between them. 
Since the number of matches within a pair varies throughout the dataset, $p \in [10000, 15000]$ random pairs are selected. 
Additionally, this means that we require much less training data, since the number of pairs available in a dataset increases according to the binomial coefficient $\binom{n}{2}$, hence resulting in $p\binom{n}{2}$ training samples.

To evaluate the pair's descriptors, pixel-wise contrastive loss is used. 
In our experiments, a margin of $m = 0.5$ was typically selected. 
This provides a good class separation, while keeping the range of values within reasonable bounds. 
Matching pairs were obtained from ground truth correspondences, whereas non-matching pairs were generated from 10 random points and averaged accordingly. 
Networks were trained using a base learning rate of 0.01 for 200 epochs, with 3 step decays of 0.1. 

One disadvantage of \acp{FCN} is the large memory requirements. 
Since fully connected layers are replaced with large 1x1 conv layers, the number of images that can be processed at any given time must be restricted. 
This becomes even more apparent when using a Siamese network. 
Due to this, images were downsampled to half-size and the batch size set to 16. 
Multiple networks were trained with varying final dimensionality output (3, 10, 32D). 
3D descriptors proved useful for visualization purposes, since they can simply be projected onto the RGB cube. 10D provides a significant improvement on 3D. 
Meanwhile, 32D typically provides a slight improvement over 10D, at the cost of less compact features. 

\vspace{-0.4cm}
\subsection{Dense feature representation}
\begin{figure}[tb!]
\vspace{-0.1cm}
\centering
\subfloat[Stereo pair]{\includegraphics[width=0.6\linewidth]{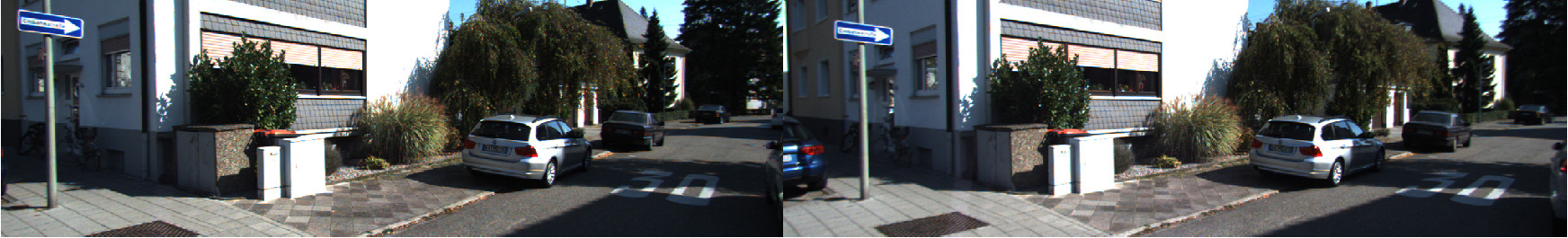}}\\
\subfloat[Base architecture]{\includegraphics[width=0.6\linewidth]{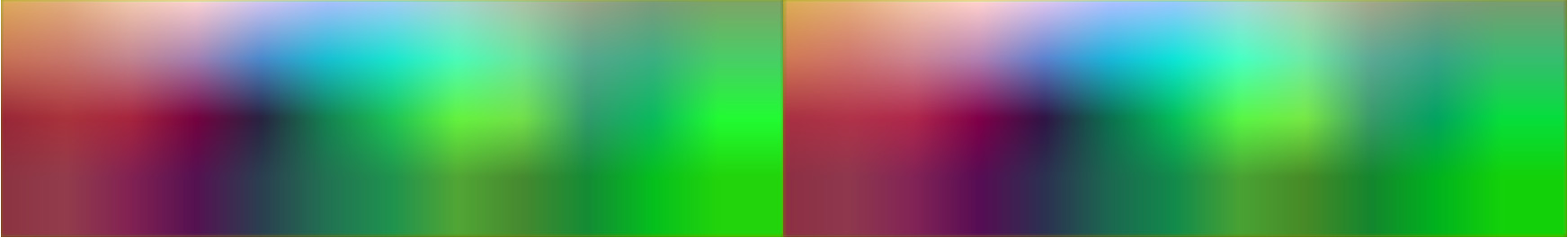} \label{fig: base_vis}}\\
\subfloat[Skip architecture]{\includegraphics[width=0.6\linewidth]{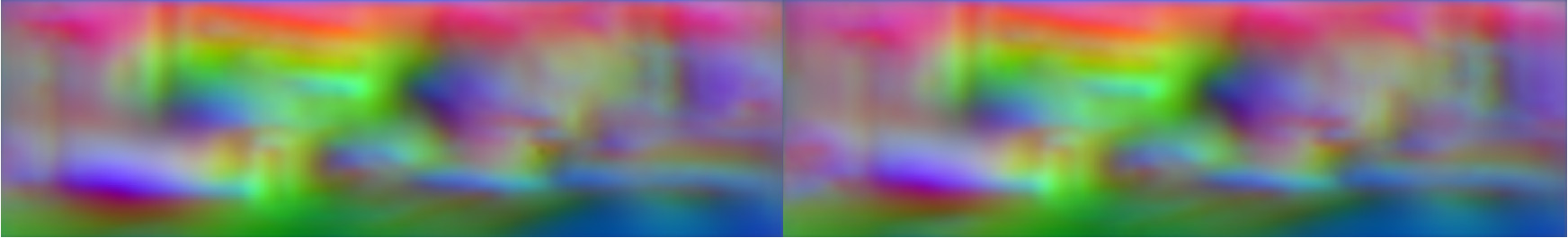} \label{fig: skip_vis}}
\vspace{-0.2cm}
\caption{Comparison between base \ac{FCU-Net}  (a) and skip connections \ac{FCU-Net}  (b) projected onto the RGB cube. Adding the skip connections allows for the combination of location information with higher level semantic meaning. This results in sharper and more discriminative features.}
\vspace{-0.4cm}
\label{fig:descs}
\end{figure}

\vspace{-0.2cm}
Multiple network architectures were used to train the feature extractors.
Comparative feature visualisations for a stereo pair, projected onto the RGB cube, are shown in Figure \ref{fig:descs}. 
Initially, a base \ac{FCU-Net}  was employed, consisting of a dowsampling stage, a bottleneck layer and a single upsampling layer. 
Class separation was still achieved, but 3D visualizations in Figure \ref{fig: base_vis} show a lack of definition and sharpness. 
Therefore, we opted for a skip connection variant (previously shown in Figure \ref{fig: fcn}). 
The upsampling is divided into several stages and merged with lower level layers of the same size. 
This further increases class separability. 
From Figure \ref{fig: skip_vis}, it can be seen that these descriptors provide a larger amount of information, with structures such as buildings and vehicles being identifiable. 

\begin{figure}[tb!]
\vspace{-0.4cm}
\centering
\subfloat[Base architecture]{\includegraphics[width=0.4\linewidth]{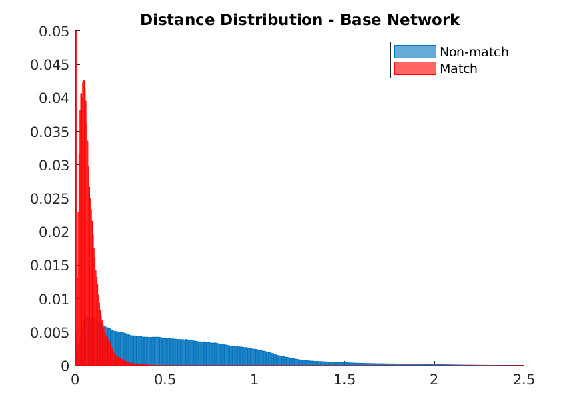} \label{fig: base_dist}}
\subfloat[Skip architecture]{\includegraphics[width=0.4\linewidth]{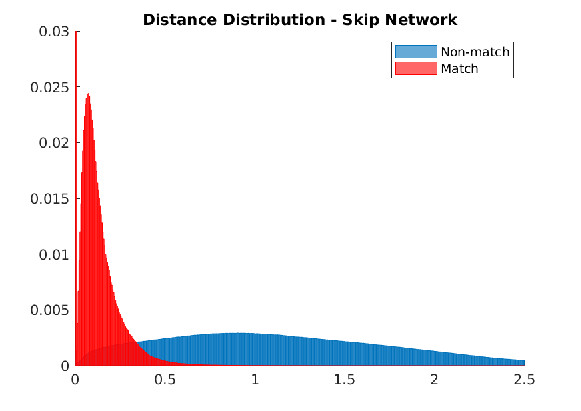} \label{fig: skip_dist}}
\vspace{-0.2cm}
\caption{Comparison between base \ac{FCU-Net}  (b) and skip connections \ac{FCU-Net}  (c) distance distributions between descriptors. In both cases, matches (red) and non-matches (blue) are seen to have significantly different distributions. However, adding the skip connections reduces the overlap between both distributions.}
\vspace{-0.2cm}
\label{fig:dists}
\end{figure}

Quantitative results are shown in Figure \ref{fig:dists}. 
This is done though the distribution of distances between previously unseen matching and non-matching features ($d$ in (\ref{eq: contrastive})). 
The distribution of match distances (red) appears very similar for both architectures, with no distances over 0.5. 
However, non-matches using the skip network (Figure \ref{fig: skip_dist}) show a larger mean distance and lower overlap than the base network (Figure \ref{fig: base_dist}). 
These results clearly show the benefits of adding skip connections within a U-Net.

In order to test the generalizing capabilities of our descriptors, we perform an evaluation using combinations of train/test datasets. 
This consists of the Kitti (K) and Apollo Scape (A) datasets. 
Table \ref{table:1} shows the results from two networks trained with Kitti and Apollo, each evaluated on both datasets. 
Here, it can be seen that, regardless of the dataset used to train, when evaluating on the same dataset, similar results are obtained. 
From here we can also deduce that the Apollo Scape dataset is harder to solve, since match distance is doubled (lower is better). 
This can also be seen in the increase in non-matching distances (higher is better) when training on Apollo. 
In general, this shows that the network has been able to learn generic and transferable features.

\begin{table}[tb!]
\centering
\begin{tabular}{||c|c|c||c|c|c||} 
 \hline
  \textit{D} & \textit{Train} & \textit{Test} & $\mu_{match}$ & $\mu_{nonmatch}$ &  $\mu_{overlap}$ \\
 \hline\hline
 3 &  &  & 0.107 & 1.156 & 0.131\\ 
 \hline
 10 & K & K & 0.103 & 1.034 & 0.138\\
 \hline
 32 &  &  & 0.103 & 1.035 & 0.139\\
 \hline\hline
 3 &  &  & 0.212 & 0.803 & 0.162\\ 
 \hline
 10 & K & A & 0.222 & 0.771 & 0.161\\
 \hline
 32 &  &  & 0.235 & 0.693 & 0.183\\
 \hline
\end{tabular}
\begin{tabular}{||c|c|c||c|c|c||} 
\hline
  \textit{D} & \textit{Train} & \textit{Test} & $\mu_{match}$ & $\mu_{nonmatch}$ & $\mu_{overlap}$ \\
  \hline\hline
  3 &  &  & 0.129 & 1.075 & 0.109\\ 
  \hline
  10 & A & K & 0.133 & 0.977 & 0.124\\
 \hline
  32 &  &  & N/A & N/A & N/A\\
  \hline\hline
  3 &  &  & 0.201 & 1.343 & 0.126\\ 
  \hline
  10 & A & A & 0.192 & 1.174 & 0.111\\
  \hline
  32 &  &  & N/A & N/A & N/A\\
  \hline
\end{tabular}
\captionsetup{width=1\textwidth}
\caption{Mean match and non-match distances for various train/test combinations on Kitti (K) and Apollo (A). In general, Apollo shows greater match distances, indicating that it is a harder dataset. This shows the generality of the learnt features.}
\vspace{-0.9cm}
\label{table:1}
\end{table}

An additional comparison between datasets and dimension combinations is performed by using \ac{LK} matching. 
Since \ac{LK} is quite a basic matcher, it provides us with information about the local uniqueness of the learnt features. 
Table \ref{table: match} shows the average distance RMSE between the matched pixel and the ground truth correspondence, along with 50th and 95th percentiles of the cumulative distribution function. 
Once again, the Kitti dataset performs better than Apollo, with a mean error of approximately 20 pixels less. 
However, one interesting thing to note from these results is the effect of descriptor dimensionality. 
While in Table \ref{table:1} all dimensions have similar average values, when performing the matching we see a significant decrease in error between 3D and 10D.\looseness=-1 

\begin{table}[tb!]
\centering
\vspace{-0.2cm}
 \begin{tabular}{||c|c|c||c|c|c||} 
 \hline
  \textit{D} & \textit{Train} & \textit{Test} & $\mu$ \textit{(pixels)} & $50th$ & $95th$ \\
 \hline\hline
 3 &  &  & 36.66 & 16.03 & 136.40\\ 
 \hline
 10 & K & K & 17.03 & 9.22 & 58.86\\
 \hline
 32 &  &  & 15.88 & 8.25 & 56.72\\
 \hline\hline
 3 &  &  & 53.85 & 30.81 & 172.24\\ 
 \hline
 10 & K & A & 35.47 & 22.02 & 112.38\\
 \hline
 32 &  &  & 34.52 & 20.10 & 111.07\\
 \hline
\end{tabular}
\begin{tabular}{||c|c|c||c|c|c||} 
 \hline
  \textit{D} & \textit{Train} & \textit{Test} & $\mu$ \textit{(pixels)} & $50th$ & $95th$ \\
 \hline\hline
 3 &  &  & 44.95 & 23.54 & 155.71\\ 
 \hline
 10 & A & K & 25.05 & 16.28 & 75.77\\
 \hline
 32 &  &  & N/A & N/A & N/A\\
 \hline\hline
 3 &  &  & 57.08 & 32.02 & 178.73\\ 
 \hline
 10 & A & A & 27.58 & 22.02 & 127.02\\
 \hline
 32 &  &  & N/A & N/A & N/A\\
 \hline
\end{tabular}
\captionsetup{width=1\textwidth}
\caption{Mean matching RMSE and percentiles for various train/test combinations on Kitti (K) and Apollo (A). Once again, Kitti generally performs better, regardless of training dataset. Additionally, the decrease in error throughout all 10D features show the uniqueness gained by increasing the latent space.}
\label{table: match}
\vspace{-0.8cm}
\end{table}

\vspace{-0.4cm}
\subsection{Deep imagination localisation}
\begin{figure}[b!]
\vspace{-0.2cm}
\centering
\begin{tabular}{lr}
\multirow{2}{*}{\subfloat[Map overview]{\includegraphics[width=0.72\linewidth]{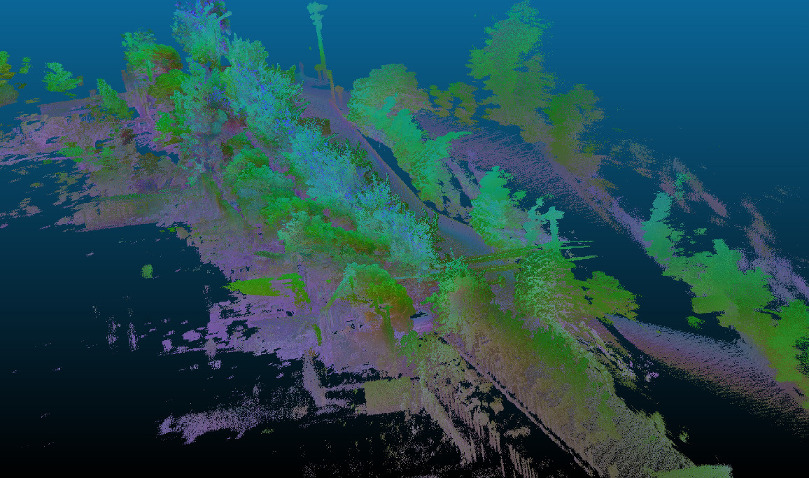} \label{fig: pc_a}}} & \\
& \subfloat[Roadsigns]{\includegraphics[width=0.285\linewidth]{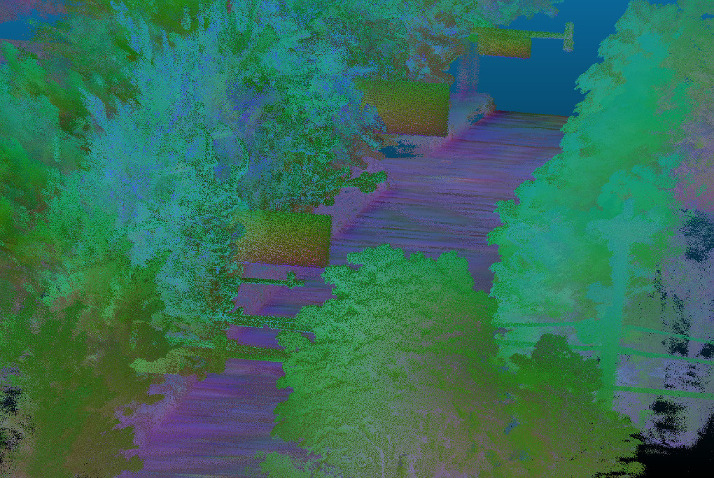} \label{fig: pc_b}} \\
 & \subfloat[Traffic light \& cables]{\includegraphics[width=0.285\linewidth]{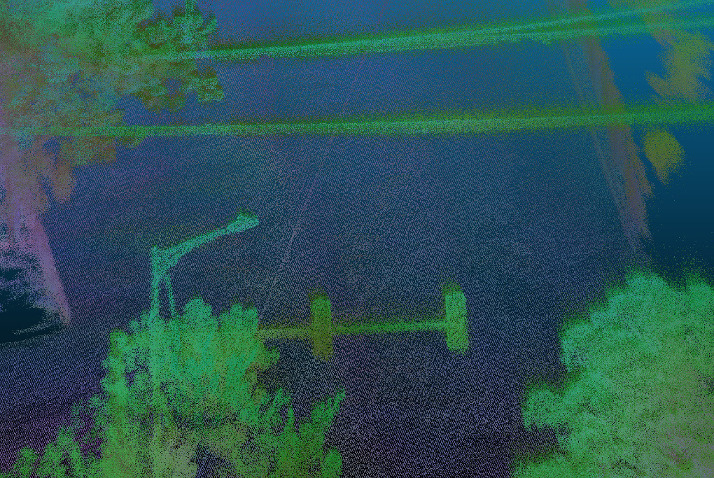} \label{fig: pc_c}}
 \end{tabular}
 \vspace{-0.2cm}
\caption{Feature embedded map visualisation. Global consistency can be seen within the trees and road overviews. Details such as the signposts also show local uniqueness.}
\vspace{-0.4cm}
\label{fig: pc}
\end{figure}

\vspace{-0.3cm}
First, we present visualizations for the generated feature embedded maps in Figure \ref{fig: pc_a}. Figures \ref{fig: pc_b} \& \ref{fig: pc_c} show some close up details of the map. These features show consistency throughout the signposts, road and overhanging cables.

\begin{figure}[tb!]
\centering
\subfloat[Novel viewpoint 1]{\includegraphics[width=0.4\linewidth]{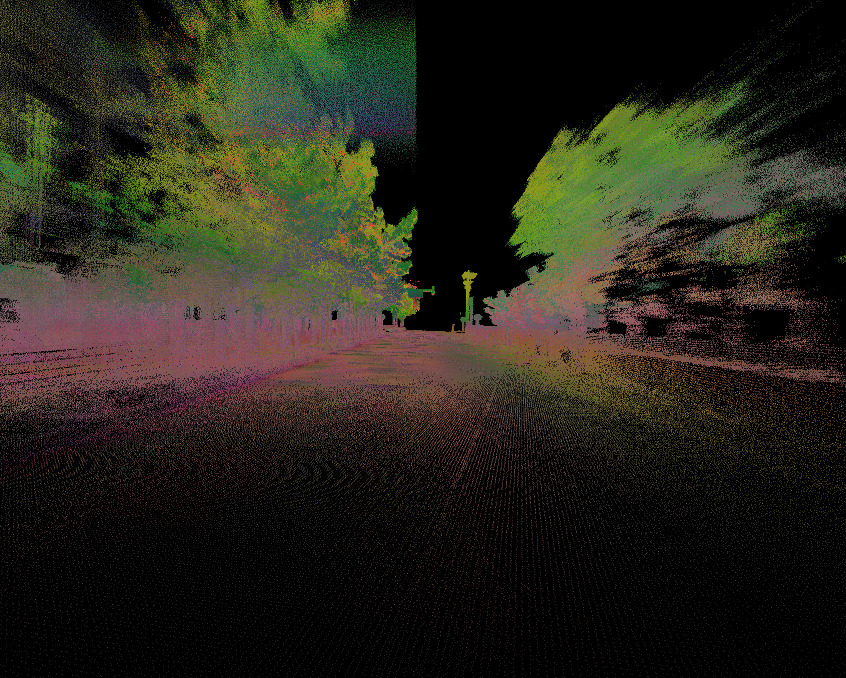} \label{fig: backview}}
\subfloat[Novel viewpoint 2]{\includegraphics[width=0.4\linewidth]{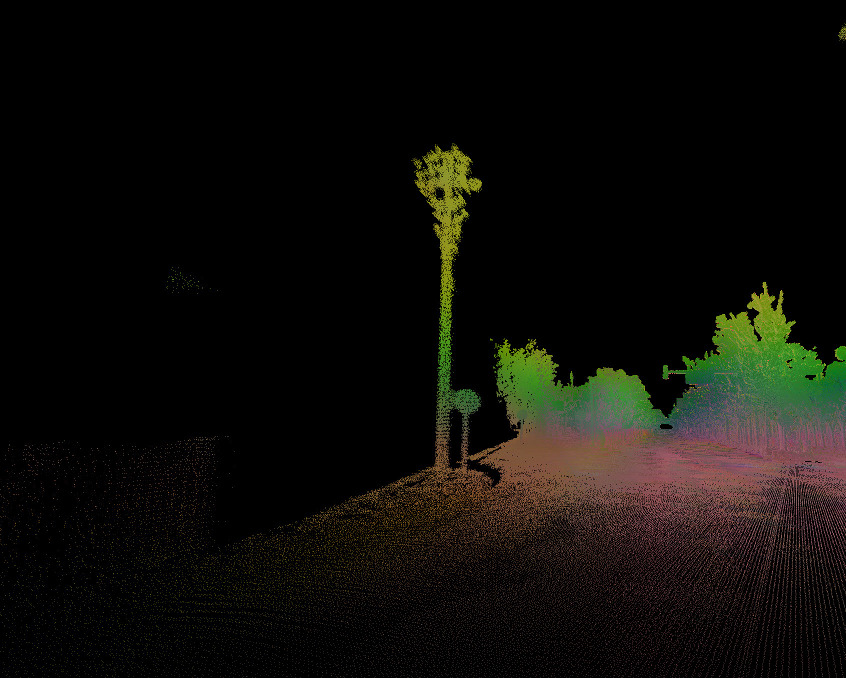}} \\
\subfloat[Imagined view]{\includegraphics[width=0.4\linewidth]{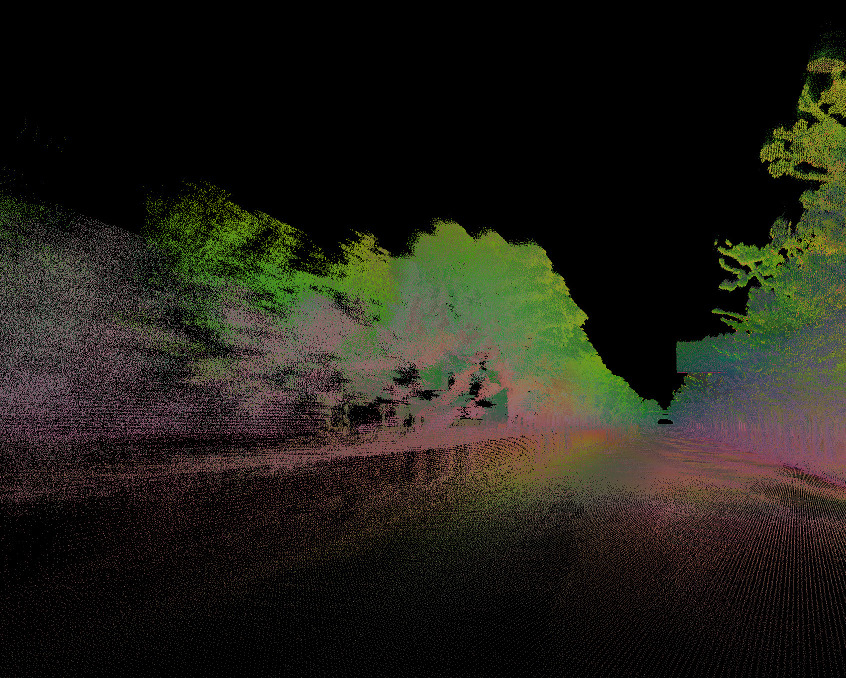} \label{fig: render}}
\subfloat[Ground truth view]{\includegraphics[width=0.4\linewidth]{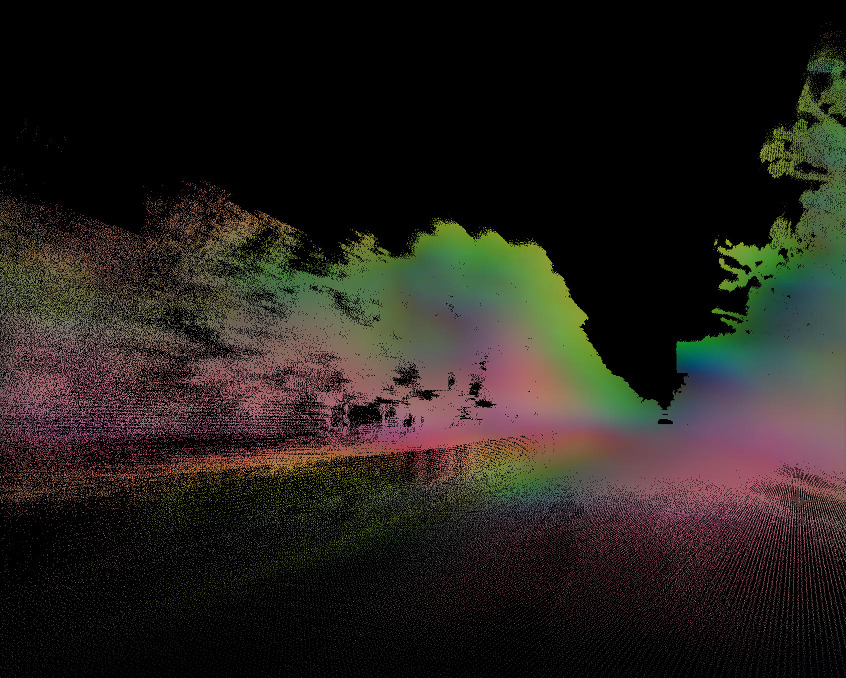} \label{fig: render_gt}}
\caption{Top row: Deep imagined viewpoints for unseen locations. Black gaps represent missing data in the map. Bottom row: Imagined view (c) and corresponding masked ground truth observation (d).}
\label{fig: imagine}
\vspace{-0.4cm}
\end{figure}

From the embedded map, we now obtain the imagined appearance from novel viewpoints, shown in the top row Figure \ref{fig: imagine}. 
By leveraging the learnt features, we can now represent the world from previously unseen positions. 
For example, Figure \ref{fig: backview} was generated facing the opposite direction of movement in the dataset. 
Black areas in these images represent missing data within the map. 
In order to provide a comparison, a view from a ground truth pose is rendered the bottom row of Figure \ref{fig: imagine}. 
Multiple similarities between the pair can be seen. 
These include the road in the horizon, the initial tree and the railing across the left side of the image. 
It also shows that having a dense feature representations enable us to compensate for sparser pointclouds with missing data, as we don't rely on particular key-points, which may or may not have been represented.

\begin{figure}[b]
\centering
\includegraphics[width=0.6\linewidth]{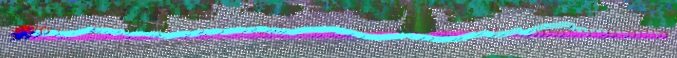}
\caption{Estimated path throughout a sequence. In general, the expected trajectory (blue) closely follows the ground truth (purple). Despite occasional drift, the system is able to correct it's position.\vspace{-0.2cm}}
\vspace{0.6cm}
\label{fig: loc}
\end{figure}

Having shown the potential of the feature embedded map and imagination systems, we proceed to use them within the localisation framework. 
A sample estimated trajectory throughout a sequence is provided in Figure \ref{fig: loc}. 
It can be seen that the agent closely follows the ground truth. 
It is also interesting to note how towards the end of the sequence, the estimated position drifts occasionally. 
However, by using the globally consistent ``Deep Imagination'' localiser, the system is able to correct it's position and effectively reset itself.

Additionally, we perform a runtime analysis of our system, shown in Table \ref{table: time}. It can be seen that the main current bottlenecks are the \ac{VO} and particle likelihood estimation. However, since \ac{VO} is performed only once per frame, this doesn't cause an issue. It is worth noting that the majority of this system has been implemented in CPU. A conversion to GPU should provide an significant speed-up to the system.

\begin{table}[tb!]
\centering
 \begin{tabular}{||c|c|c||} 
 \hline
  Subsystem & Result & Time\\
  \hline\hline
  Feature Extraction & $F$ & $1.93 \pm 0.23$ ms/frame\\
  \hline
  Visual Odometry & $\Delta\boldsymbol{\boldsymbol{\omega}}$  &  $201 \pm 100$ ms/frame\\
  \hline
  Particle Initialization & $S_0$ & $4.28 \pm 0.09$ us/particle (once only)\\
  \hline
  Particle Resampling & $S_i$ & $3.40 \pm 0.33$ us/particle\\
  \hline
  Particle Weights & $P(I_{t}|\boldsymbol{\omega}_{t},\mathcal{M})$ & $7.36 \pm 0.25$ ms/particle \\
  \hline
  Final Location Estimate &  $\tilde{\boldsymbol{s}}$& $3.27 \pm 0.79$ ms/particle\\
  \hline
\end{tabular}
\captionsetup{width=0.8\textwidth}
\caption{Time taken for each of the subsystems within the localisation implementation.}
\label{table: time}
\vspace{-0.8cm}
\end{table}

\vspace{-0.6cm}
\section{Conclusions \& future work}
\label{sec: conc}
\vspace{-0.4cm}
We have presented a novel method for localisation within a known environment. This was achieved through a ``Deep Imagination'' localiser capable of generating views from any position in an existing feature embedded 3D map. In order to construct the embedded map, a deep dense feature extractor was trained in the form of a Siamese \ac{FCU-Net}. By learning generic features, training is limited to a single initial network. These features can then be applied to new unseen environments, requiring little to no additional training data. In turn, this means that we are able to build a representation of a new environment and ``imagine'' what it looks like from any given position after a single visitation.

From the presented results in Tables \ref{table:1} \& \ref{table: match}, it can be seen that our feature descriptors show a good level of generality and can be used with previously unseen images and datasets. It is worth noting that this was achieved using only a small fraction of both datasets. By using a larger amount of data, including pairs with fewer or more complicated matches, it should be possible to further improve these results. Additionally, the \ac{LK} matching results show that by increasing the latent space available we can both globally and locally discriminative features. This opens possibilities to a new method for \ac{VO} estimation. Subjectively, the 3D descriptor visualizations give some insight into what the network has learnt. This opens the door to further interpretability studies.  

As future work, we plan to incorporate the generic dense features into the \ac{VO} pipeline. This would allow for dense matching of the images, without relying on key-point feature detection. Additionally, the ``Deep Imagination'' localiser may be improved by introducing a more sophisticated pose likelihood calculation (P\textit{n}P) or additional motion models (non-holonomic constraints). Finally, we are interested in exploring DSAC \cite{Brachmann} and it's applications, including it's potential use within the Siamese \ac{FCU-Net} training pipeline.

\subsection*{Acknowledgements}
\vspace{-0.3cm}
This work was funded by the EPSRC under grant agreements (EP/R512217/1) and (EP/R03298X/1) and Innovate UK Autonomous Valet Parking Project (Grant No 104273). 
We would also like to thank NVIDIA Corporation for their Titan Xp GPU grant.
\vspace{-0.6cm}

\bibliographystyle{splncs}
\bibliography{ApolloScape}
\end{document}